\documentclass{Interspeech}
\usepackage{url}

\interspeechcameraready

\title{What Do Humans Hear When Interacting? Experiments on Selective Listening for Evaluating ASR of Spoken Dialogue Systems}

\author[affiliation={1,2}]{Kiyotada}{Mori}
\author[affiliation={2,1}]{Seiya}{Kawano}
\author[affiliation={2,3}]{Chaoran}{Liu}
\author[affiliation={2}]{Carlos}{Toshinori Ishi}
\author[affiliation={2}]{Angel}{Fernando Garcia Contreras}
\author[affiliation={4,2,1}]{Koichiro}{Yoshino}

\affiliation{}{Nara Institute of Science and Technology}{Japan}
\affiliation{Guardian Robot Project}{RIKEN}{Japan}
\affiliation{}{National Institute of Informatics}{Japan}
\affiliation{}{Tokyo Institute of Technology}{Japan}

\email{mori.kiyotada.mh5@naist.ac.jp,
seiya.kawano@riken.jp,
cliu@nii.ac.jp,
carlos.ishi@riken.jp,
angel.garciacontreras@riken.jp,
yoshino.k.ai@m.titech.ac.jp}

\usepackage{comment}

\begin{document}

\maketitle
\begin{abstract}
Spoken dialogue systems (SDSs) utilize automatic speech recognition (ASR) at the front end of their pipeline. The role of ASR in SDSs is to recognize information in user speech related to response generation appropriately. Examining selective listening of humans, which refers to the ability to focus on and listen to important parts of a conversation during the speech, will enable us to identify the ASR capabilities required for SDSs and evaluate them. In this study, we experimentally confirmed selective listening when humans generate dialogue responses by comparing human transcriptions for generating dialogue responses and reference transcriptions. Based on our experimental results, we discuss the possibility of a new ASR evaluation method that leverages human selective listening, which can identify the gap between transcription ability between ASR systems and humans.
\end{abstract}

\section{Introduction}
\label{sec:intro}

Spoken dialogue systems (SDSs) listen to, understand, and respond to the user's speech. 
In such systems, automatic speech recognition (ASR) performance strongly affects the accuracy of subsequent tasks such as dialogue response generation. 
The accuracy of ASR systems is currently very high by benefiting from the development of deep learning. 
For example, Whisper~\cite{radford2023robust} was trained on 680,000 hours of multilingual speech data and achieved very high performance, comparable to human professional transcribers, in evaluating word error rate (WER).

However, when using SDSs, dialogue response failures are still caused by ASR errors~\cite{chang2023dialog,wang2024retrieval}. 
WER evaluates each word equally. 
This property is particularly useful in dictation, where each word needs to be evaluated equally. 
On the other hand, when the importance of words needs to be considered, such as in a dialogue, this approach does not necessarily guarantee accuracy comparable to human accuracy~\cite{he2011word,kafle2017evaluating,rugayan23perceptual}. 
Existing speech recognition evaluation research for SDSs, such as keyword error rate \cite{park2008empirical} and concept error rate \cite{boros1996towards} require human references in order to appropriately evaluate word importance. Moreover, to specify each word importance, clear tasks such as restaurant or bus reservation is required. Thus, these methods are applicable only to specific cases within task-oriented dialogue where human reference can be used.
Our research aims to find a solution to this problem by observing human listening behavior during a dialogue.

One phenomenon that suggests a method for appropriately measuring the ASR capabilities required for SDSs is selective listening~\cite{hari1989selective} performed by humans. 
Selective listening is the subconscious ability of humans to focus on important parts of a conversation and ignore irrelevant sounds. 
This ability allows humans to converse with a specific person, even in a noisy environment. 
The cocktail party effect is a typical example of this phenomenon~\cite{arons1992review}. 
Selective listening has been verified through psychological experiments~\cite{woldorff1991modulation,woods1984event}, and there is strong evidence for its existence and function.

In particular, for SDSs to function correctly in noisy environments, it is necessary to evaluate whether ASR-based transcriptions achieve a performance comparable to such selective listening by humans. 
However, to develop such evaluation metrics, it is first necessary to correctly understand the behavior of humans when they perform dialogue responses in noisy environments. 
Existing research on the transcription task has confirmed that humans and ASR models are more likely to make errors with content words than function words~\cite{bell2009predictability,mansfield2021revisiting}. However, humans' listening behavior in tasks requiring accurate transcription may be different from that in tasks requiring dialogue response generation. Conducting experiments to understand what humans listen to during dialogue responses is necessary.

In this study, we solve this problem by combining the dialogue response generation task and the transcription task in a noisy environment to clarify humans' selective listening behavior during dialogue responses. 
In our experiment, we informed the subjects that they would be performing dialogue response generation. 
We presented them with an image indicating a conversation scene and a relatively short speech corresponding to the scene only once. 
After the response generation, the subjects were asked to transcribe the speech they heard based on their memory immediately, a task that had not been instructed in advance. 
By experimenting this way, we clarified the degree of attention paid to each word in the dialogue response, which is different from ordinary transcription tasks.

The analysis of the transcripts based on human recall revealed that humans pay more selective attention to different parts of speech during dialogue than in standard transcription tasks. 
Specifically, we observed completely different trends, especially for content words and function words. 
Based on these trends, we compared human transcription in dialogue with ASR transcription, and the results suggested that WER-based evaluation may not accurately evaluate the content used by humans to generate dialogue responses.

Our contribution is that by focusing on human selective listening, it has been shown that the conventional ASR evaluation method for SDSs based on WER is not necessarily optimal. These results will serve as a bridge between psychological experiments on selective listening and ASR evaluation methods for SDSs. 
They may be useful in the development of new ASR evaluation methods for SDSs in the future.

\section{Experimental Procedure}
\label{sec:typestyle}
We conducted an experiment to analyze selective listening in humans when generating dialogue responses. 
The experiment procedure follows.
We used a portion of a visually grounded first-person dialogue (VFD)~\cite{kamezawa2020visually} dataset. 
The VFD contains first-person images and utterance texts corresponding to the images, and four or five examples of Japanese verbal responses. We refer to the verbal responses as $r_{\mathrm{VFD}}$. Subjects were recruited to generate subsequent dialogue responses based on the sets of query images and utterance texts readout, short speeches\footnote{The experiment was conducted from September 4 to September 8, 2023, using CrowdWorks (https://crowdworks.jp), a crowdsourcing service, in conjunction with GitHub Pages.}. We transcribed the audio stimuli accurately. We refer to the gold transcription as $q_{\mathrm{VFD}}$.
297 Japanese-speaking subjects participated.

\begin{figure}[t]
\begin{minipage}[b]{1\columnwidth}
    \centering
\includegraphics[width=0.75\columnwidth]{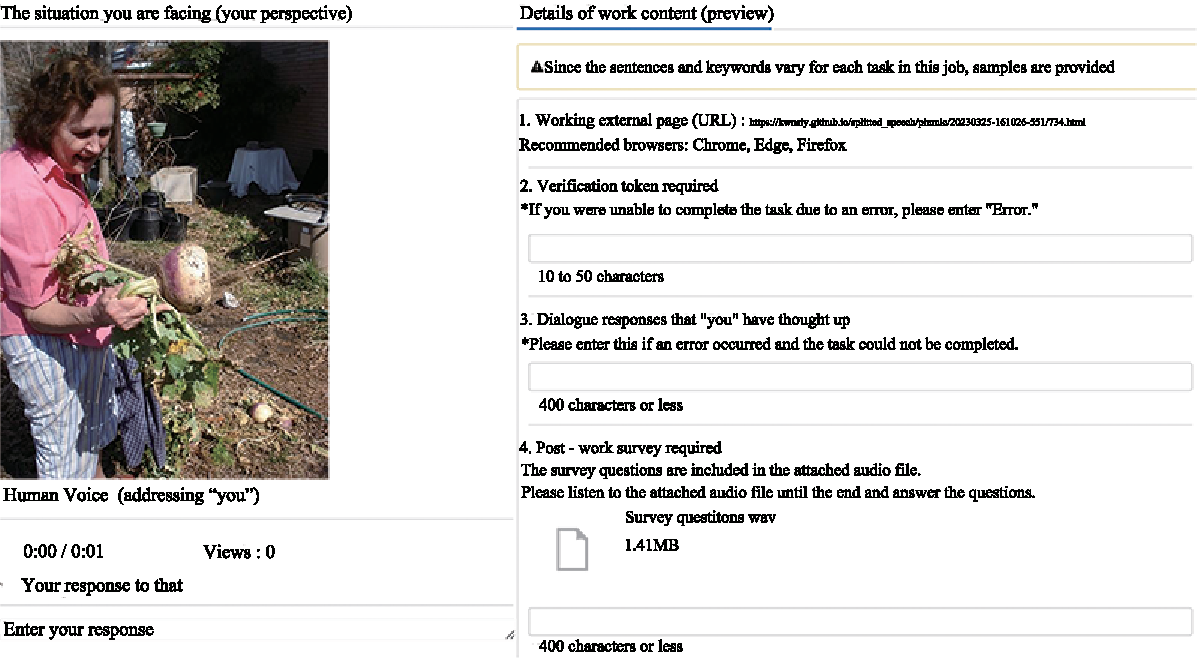}
\end{minipage}
\caption{Query Image \& Answer Form}
\label{fig:form}
\end{figure}
The subjects were directed to the web page \footnote{\url{https://kwnsiy.github.io/splitted_speech/pinmic/20230325-161026-551/400.html}}
where they engaged in dialogue response generation by following an answer form shown in Figure~\ref{fig:form}.
The subjects listened to the test audio to verify that their devices were functioning properly and that they could clearly hear the audio. 

For the first task, while viewing the query image, the subjects listened to the audio stimuli only once and wrote a dialogue response to them. 
They focused on providing meaningful responses that encouraged further dialogue. 
They were required to avoid overly generic, bland remarks like ``I see.'' 
For the second and hidden task, subjects move to the next webpage and are instructed to open an audio file of the post-work survey, which involves an unexpected task: a rigorous transcription of the audio stimuli\footnote{The given instruction follows: ``What was the speech in the audio file, which you generate dialogue responses to in this work? Remember the audio as accurately as possible, and type them into the text box. When your memory is fuzzy or you are unsure what you heard, just enter anything related to the audio: a word, a phrase, a string, or a topic. Please do not return to the previous page to listen to the audio again.''}. 
We hoped that this procedure would show results indicating selective listening when humans listen for dialogue responses. 
Note that we can collect only one sample from one subject in this way.

Examples of the data obtained in the experiment are shown in Table~\ref{tb:example}. 
\begin{table}[t]
  \caption{Example Data Used in the Experiment}
  \centering
  \begin{tabular}{ll}
    \hline
    Original&Japanese\\
    \hline
    $q_{\mathrm{VFD}}$&{\footnotesize \includegraphics[height=0.9em]{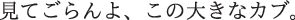}}\\
    $q_{\mathrm{subj}}$&{\footnotesize \includegraphics[height=0.9em]{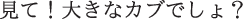}}\\
    $r_{\mathrm{VFD}}$&{\footnotesize \includegraphics[height=0.9em]{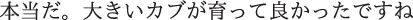}}\\
    $r_{\mathrm{subj}}$&{\footnotesize \includegraphics[height=0.9em]{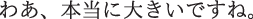}}\\
    \hline
    Translation&English\\
    \hline
    $q_{\mathrm{VFD}}$&{\footnotesize Look at this big turnip.}\\
    $q_{\mathrm{subj}}$&{\footnotesize Look! Big turnips, right?} \\
    $r_{\mathrm{VFD}}$&{\footnotesize True. I'm glad you grew a big turnip.}\\
    $r_{\mathrm{subj}}$&{\footnotesize Wow, it's huge.} \\
    \hline
  \end{tabular}
  \label{tb:example}
\end{table}
We define items for analysis as follows: 
\begin{itemize}
\setlength{\itemsep}{0cm} %
\item $q_{\mathrm{subj}}$: Transcribed audio stimuli by a subject
\item $r_{\mathrm{subj}}$: Generated response to audio stimuli by a subject
\end{itemize}

$q_{\mathrm{VFD}}$ and $q_{\mathrm{subj}}$ were used to confirm selective listening. As shown in example, $q_{VFD}$ and $q_{subj}$ have the same meaning but the detailed parts are different. 
$r_{\mathrm{VFD}}$ and $r_{\mathrm{subj}}$ were used to confirm that human subjects generated qualified dialogue responses in our task setting; they focused on response generation appropriately, in other words.

Although there were initially 297 samples, due to issues in the
experimental procedure, only 274 samples matched both the audio stimulus and corresponding dialogue responses. 
We modified the notation of $q_{\mathrm{subj}}$ in some cases. 
For example, some subject put their transcription as ``He said hello,'' it was revised to ``Hello''. 
Some samples that failed to fit the question's intention were deleted, such as the subject tried rewriting his/her dialogue responses when they were required to put the transcription. 
After the reformatting and revising, out of 274 samples, 246 available samples remained for the analysis. 

\begin{figure}[t]
\begin{minipage}[b]{1\columnwidth}
    \centering
    \includegraphics[width=0.75\columnwidth]{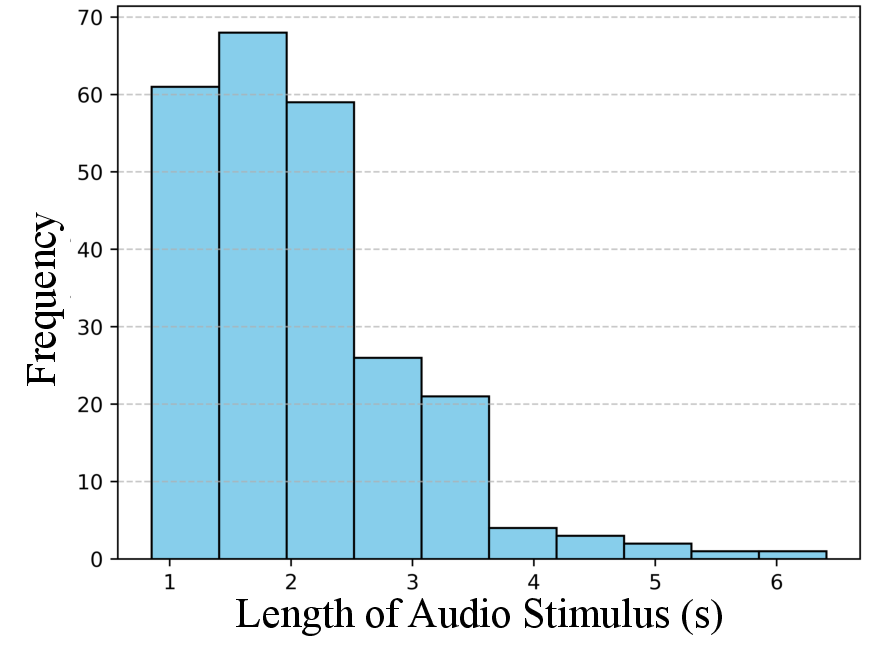}  
    \caption{Speech Duration Distribution}
    \label{fig:duration}
\end{minipage}
\end{figure}

We used VFD readings as audio stimuli recorded in noisy areas of a shopping mall. 
The 246 audio stimuli used in our experiment had an average duration of 2.08 seconds and a variance of 0.79, with most audio stimuli being less than 4 seconds, as shown in Figure~\ref{fig:duration}. 
The duration was short enough to be fully reproduced if attention was directed to them, because humans have the ability to transcribe such audio accurately if the task is transcription \cite{xiong2016achieving}. 

The audio was recorded with a 16-channel microphone array, and we mixed them into a single channel to create the audio stimulus for human subjects.
For ASR systems used for the following experiments, the audio stimulus was enhanced with sound source-positioned beamforming~\cite{ishi2015speech}. 

\section{Evaluation Methods}
We analyzed human selective listening by comparing human transcriptions obtained from the experiment with transcriptions from the ASR model. 
Two ASR models, Whisper base and Whisper large-v2, transcribed the audio stimulus to compare with the human transcriptions.
The ASR models also transcribed the audio stimulus with an additional visual-conditioned prompt for the query images using PromptingWhisper~\cite{peng2023prompting} in order to match the input conditions for humans and ASR models as much as possible. Google Cloud Translate v2 translated PromptingWhisper's visual conditional prompts from English to Japanese to provide Japanese vocabulary for the model. 

We calculated WER, Character Error Rate (CER), which is a character-based version of WER, and the semantic similarity of these transcriptions to $q_{\mathrm{VFD}}$ using multilingual Sentence BERT (S-BERT), stsb-xlm-r-multilingual \cite{reimers-2019-sentence-bert}.
Since transcription behavior is expected to differ between content words and function words, we prepare evaluation metrics as follows: ``WER CONT'' for indicating WER of content words, ``WER FUNC'' for indicating WER of function words, ``CER CONT'' for indicating CER of content words, and ``CER FUNC'' for indicating CER of function words. 
For comparing the recognition accuracy of function words and content words, only the function words or content words were extracted after removing spaces from $q_{\mathrm{subj}}$. 
Content words were defined as nouns, verbs, adjectives, adverbs, and proper nouns. 
Function words were defined as particles, auxiliary verbs, prepositions, conjunctions, pronouns, numerals, subordinating conjunctions, determiners, and interjections. 
We used a morphological analysis engine GiNZA \cite{Matsuda2021} to extract the function and content words. 
In Japanese, the same word may be written using different kana or kanji. 
To address this issue, we converted all kanji in the transcriptions to hiragana in accordance with previous research on Japanese speech recognition. 
This process may make it difficult to evaluate homonyms, but we did not observe any such cases in the audio stimuli used in this study.
We converted each character to hiragana using Pykakasi \cite{Miura2020} for the evaluation.

To confirm that humans appropriately solved the dialogue response tasks in the experiment, we evaluated the dialogue responses of $r_{\mathrm{VFD}}$ and $r_{\mathrm{subj}}$ with Athena-RR \cite{harrison2023transformer}, a response ranking model.
Athena-RR was trained to binary classify whether a given dialogue response is a candidate response for a given context. 
In this experiment, $q_{\mathrm{VFD}}$ was input as the context. 
The candidate dialogue responses were $r_{\mathrm{VFD}}$ and $r_{\mathrm{subj}}$. 
Since Athena-RR only supports English, responses were translated from Japanese to English by Google Cloud Translate v2.

\section{Result}
\label{sec:majhead}
Table~\ref{tb:WER} shows the results of our experiments.
We confirmed that humans exhibit selective listening to generate dialogue responses by analyzing the WER and semantic similarity of $q_{\mathrm{VFD}}$ and $q_{\mathrm{subj}}$. 
If the audio stimulus ($q_{\mathrm{VFD}}$) does not contain content words or function words, we eliminated them from the calculation of WER FUNC and WER CONT. 

When comparing Base~A+V and Human~A+V, we found that, despite Base~A+V being superior in WER, the human transcription had a higher semantic similarity with the reference on S-BERT. Large-A+V also shows a similar trend to Base~A+V.
This means that humans focus more on the semantic aspects of speech in dialogue response generation, reproducing transcriptions that are more semantically similar. 
In addition, we can see that Human~A+V transcribes content words more accurately than Base~A+V but is less accurate in transcribing function words. 
This suggests that when humans generate dialogue responses, they exclude function words that do not significantly contribute to content understanding from their attention and place more importance on content words.

We used a two-tailed paired t-test to determine whether there is a significant difference between recognition accuracy of the function and contents words, with a significance level of 5\%. 
The null hypothesis is that no difference exists between the recognition accuracy of the function and content words. The results of significance tests are shown in Table~\ref{tb:WER-ttest}. 
The null hypothesis was rejected in CER of Human~A+V, but not rejected in WER of Human~A+V. 
The results suggested that humans pay more attention to content words captured by CERs. 
PromptingWhisper's visual speech recognition results also show the same trend as the results of the speech-only input.

\begin{table}[t]
\caption{Mean WERs on Content and Function Words and Sentence BERT Score\\\centerline{(A: audio, V: visual)}}
\label{tb:WER}
\centering
  \begin{tabular}{llllll}
    \hline
    Model&Human&Base&Large&Base&Large\\
    Modality&A+V&A&A&A+V&A+V\\
    \hline
    WER &0.41&0.40&0.12&0.36&0.17\\
    WER FUNC &0.48&0.29&0.14&0.28&0.17\\
    WER CONT &0.40&0.51&0.14&0.44&0.18\\
    \hline
    CER & 0.32&0.22&0.06&0.27&0.16\\ 
    CER FUNC & 0.44&0.29&0.10&0.24&0.14\\
    CER CONT &0.35&0.31&0.09&0.32&0.18\\
    \hline
    S-BERT & 0.87&0.76&0.94&0.77&0.93\\ 
    \hline
  \end{tabular}
\end{table}
\begin{table}[t]
  \caption{Paired t-test for WERs on Content and Function Word \\\centerline{(A: audio, V: visual, $^{*}$: $p<0.10$, $^{**}$: $p<0.05$)}}
  \centering
  \label{tb:WER-ttest}
  \begin{tabular}{llllll}
    \hline
    Model&Human&Base&Large&Base&Large\\
    Modality&A+V&A&A&A+V&A+V\\
    \hline
    p (WER)&$\mathbf{0.06^*}$&$\mathbf{0.00^*}$&0.42&$\mathbf{0.00^{**}}$&0.52\\
    p (CER)&$\mathbf{0.02^{**}}$&0.77&0.65&$\mathbf{0.07^*}$&0.93\\ 
    \hline
  \end{tabular}
  \label{tb:fugafuga}
\end{table}

The average ratings of the dialogue responses of $r_{\mathrm{VFD}}$ and $r_{\mathrm{subj}}$ by Athena-RR were 0.56 and 0.64. 
The null hypothesis was that $r_{\mathrm{subj}}$ has a lower rating than $r_{\mathrm{VFD}}$, and a one-tailed U-test was conducted, with a significance level of 5\%. 
As a result, the null hypothesis was not rejected. 
It probably means that the subjects can respond to the same degree as the response examples.

Humans can respond despite having higher WER and semantic similarity than the ASR model. The WER of the content words is significantly lower than that of the function words. Therefore, we confirmed that humans focus on listening to content words to generate dialogue responses. In other words, the results suggest that a WER that does not leverage selective listening is an inappropriate evaluation method for speech recognition systems to be used for dialogue systems.

\section{Discussion of New Metrics for Evaluating ASR in SDSs}
From the experiments, we have learned that human listening during speech interaction tends to focus on content words. 
It is difficult to adequately measure the ASR performance for SDSs using existing WER alone.
We compared $q_{\mathrm{subj}}$ with $q_{\mathrm{VFD}}$ to identify trends that humans were unable to transcribe, implicitly judging it as unimportant, and estimated the importance of each part of speech (POS) using the following method.

Multiple regression analysis was done with the probabilistic gradient descent method to analyze selective listening to different POS. 
An explanatory variable was the number of each POS of $q_{\mathrm{VFD}}$. 
A target variable was the sum of the number of each POS of $q_{\mathrm{subj}}$.  
The partial regression coefficients for the POS are expected to be small, a state that tends to avoid attention when generating dialogue responses. 
Conversely, when the partial regression coefficients for the POS are expected to be significant, this state tends to attract attention. 
The calculation for the multiple regression analysis is shown in Equation (\ref{eqn:regression}):
\begin{align}
Y = \beta_0 * \text{NOUN} + \beta_1 * \text{PROPN} + \beta_2 * \text{VERB} + \cdots \label{eqn:regression}
\end{align}

The target variable $Y$ is the sum of the number of POSs of $q_{\mathrm{subj}}$, and $\beta_n$ is the partial regression coefficient. 
Explanatory variables $\text{NOUN}$ and $\text{PROPN}$ represent the number of POSs of $q_{\mathrm{VFD}}$. 
The same is true for other POSs. 
``Symbol'' POS was excluded from the explanatory variables. 
The intercept was fixed at $0$ to clarify the meaning of the partial regression coefficients.  

Table~\ref{tb:regression-cross} indicate a mean absolute error (MAE) and the coefficient of determination ($R^2$) for the test data of the model optimized with $q_{\mathrm{subj}}$ and $q_{\mathrm{VFD}}$, respectively. 
 $MAE_{\mathrm{Uni}}$ and $R^2_{\mathrm{Uni}}$ represent the MAE and the $R^2$ for the test data of the model where all the POS weights are 1. The $MAE_{\mathrm{Optim}}$ and $R^2_{\mathrm{Optim}}$ represent the same methods for the optimized model.
The model optimized by multiple regression for $q_{\mathrm{subj}}$ and $q_{\mathrm{VFD}}$ has a smaller MAE than the model where every POS weight is 1. It indicates that the model obtained by multiple regression is more adaptable to unknown data than the model with equivalent weights for every POS.

The coefficients for each POS obtained by multiple regression are shown in Table~\ref{tb:attention-POS}. 
The results show that humans pay more attention to content words, such as nouns, verbs, and adjectives, and less attention to function words, such as particles and auxiliary verbs. 
The weights in Table~\ref{tb:attention-POS} were the coefficients with the lowest MAE in the 5-fold cross-validation. 

The weights of each POS obtained were then used to define Human-WWER (H-WWER) as a new Weighted WER (WWER)~\cite{nanjo2005minimum} that leverages human selective listening, using the weights of each POS obtained in the regression. 
The weight of each operation in the minimum edit distance for WER varies depending on the POS of the word being edited.

Table~\ref{tb:WWER} shows the results of the H-WWER evaluation of the human (Human~A+V), Whisper base (Base~A and Base~A+V), and Whisper large-v2 (Large~A and Large~A+V) transcriptions. 
Unlike WER, the H-WWERs of Human transcription are lower than the H-WWERs of Whisper base transcription.
Of course, the weight parameters used for Table~\ref{tb:WWER} are determined by the test data; such a comparison is unfair.
However, this result suggests that by defining speech recognition evaluation methods that leverage human selective listening, perhaps we can evaluate whether an ASR model is transcribing the necessary content for dialogue responses.
\begin{center}
\begin{table}[t]
  \caption{5-fold Cross Validation for Multiple Regression}
  \centering
  \begin{tabular}{lllll}
    \hline &$MAE_{\mathrm{Optim}}$&$MAE_{\mathrm{Uni}}$&$R^2_{\mathrm{Optim}}$&$R^2_{\mathrm{Uni}}$\\
    \hline
    Mean&\textbf{1.53}&1.70&\textbf{0.70}&0.55\\
    \hline
  \end{tabular}
  \label{tb:regression-cross}
\end{table}
\end{center}
\vspace{-1cm}  
\begin{table}[t]
  \caption{Human Attention for Part-of-speech}
  \centering
  \begin{tabular}{ll|ll}
    \hline
    POS&Weight&POS&Weight\\
    \hline
    \textbf{ADJ}&1.41&AUX&0.85\\
    \textbf{ADV}&1.38&NUM&0.21\\
    \textbf{VERB}&1.18&INTJ&0.21\\
    \textbf{NOUN}&1.12&ADP&0.13\\
    PROPN&0.19&PRON&0.84\\
    PART&1.51&PUNCT&0.50\\
    DET&1.12&CCONJ&-0.27\\
    SCONJ&0.90&SYM&0.00\\
    \hline
  \end{tabular}
  \label{tb:attention-POS}
\end{table}
\begin{table}[t]
  \caption{H-WWER on Transcriptions (A: audio, V: visual)}
  \centering
  \begin{tabular}{llllll}
    \hline
  Model&Human&Base&Large&Base&Large\\
    Modality&A+V&A&A&A+V&A+V\\
    \hline
    H-WWER&\textbf{0.48}&0.56&0.28&0.57&0.33\\
    H-WCER&\textbf{0.38}&0.41&0.20&0.44&0.26\\
    \hline
  \end{tabular}
 \label{tb:WWER}
\end{table}

\section{Related Works}
\label{sec:format}

Speech recognition evaluation methods instead of WER, Semantic Distances~\cite{kim21e_interspeech}, which are based on semantic similarity, have been proposed. 
Moreover, Sasindran~et~al. proposed $H_{eval}$~\cite{sasindran2023h}, a hybrid speech recognition evaluation method that uses WER and semantic similarity advantages. 
$H_{eval}$ detects important and unimportant words by semantic similarity and derives the minimum edit distance for each.
Our experiment supports the approaches of these previous studies from the perspective of human behavior during dialogue.

Millet~et~al.~\cite{millet20_interspeech} confirmed that supervised ASR models in phoneme identification differed significantly from human perceptual space. 
A typical supervised fine-tuned ASR model such as Whisper and human speech recognition errors will likely differ significantly, as confirmed in our experiment.

Wikman~et~al.~\cite{wikman2022brain} used functional magnetic resonance imaging to analyze differences in brain activity between shadowing and listening using audiovisualized dialogue. 
Their results showed that an identical frontal network was activated, although its peak activation location differed. 
The results of this experiment suggest that different kinds of selective listening may occur in dialogues when responding and listening attentively from a neuroscience perspective.
In our experiment, we confirmed this trend of selective listening during dialogue.

\section{Conclusion}
\label{sec:page}
This paper conducted experiments on selective listening during human dialogue response generation. 
We designed a human subject experiment to confirm our hypotheses that human has different trends to ASR systems in response generation and conducted a large-scale evaluation with 297 subjects.
Our results confirmed that humans exhibit selective listening for sounds during responses, and human transcriptions after the response generation have different trends to transcriptions of existing ASR systems.
This result also suggests that WER is not necessarily the best evaluation metric for ASR modules in SDSs, from a novel perspective, human perception.
Based on our analysis, we proposed H-WWER, which leverages the selective listening of humans to POS, and has the potential to more appropriately evaluate the accuracy of ASR systems in terms of recognizing what is necessary for dialogue responses than WER.

Future work will first address how the human transcription used in experiments differs from transcriptions made immediately before responses. This difference stems from the experimental limitation that humans cannot transcribe and respond simultaneously. 
One way to address this issue is using biological signals such as electroencephalography~\cite{xie2024eeg}.
To prevent such a problem caused by human working memory~\cite{baddeley1992working}, we can also refine the experimental process in the future.

Although our study identified an average trend of selective listening for each POS, the degree of attention given to different parts of speech will likely vary depending on the language. 
A more detailed analysis is necessary based on actual dialogue contents across languages.
\clearpage
\section{Acknowledgement}
A part of this work was supported by JST MOONSHOT Grant Number JPMJMS2236 and JSPS KAKENHI 22K17958.
\bibliographystyle{IEEEtran}
\bibliography{mybib}

\end{document}